# SYNERGISTIC FOUNDATION MODELS FOR SEMI-SUPERVISED FETAL CARDIAC ULTRASOUND ANALYSIS: SAM-MED2D BOUNDARY REFINEMENT AND DINOV3 SEMANTIC ENHANCEMENT

*Tonghao Zhuang*[1*], *Shanglong Hu*[2*], *Yongsheng Luo*[3], *Zhiqi Zhang*[4], *Yu Li*[5†]

[1,2,3,4,5]Zhuhai College of Science and Technology, Zhuhai, China
[1*]*tonghaozhuang39@163.com*, [2*]*shanglonghu@stu.zcst.edu.cn*, [3]*vincent@zcst.edu.cn*,
[4]*2282625344@qq.com*, [5†]*jluzhliyu@zcst.edu.cn*

## ABSTRACT

We present a semi-supervised framework for joint segmentation and classification of fetal cardiac ultrasound images. Built upon the EchoCare multi-task backbone, our method integrates SAM-Med2D for boundary refinement and leverages DINOv3 to enhance pseudo-label quality. We introduce view-specific hard masking along with a two-stage optimization strategy: an *EMA phase* to consolidate segmentation capabilities, followed by a *Classification Fine-Tuning phase* that freezes segmentation parameters and resets the classification head to recover classification performance without compromising segmentation gains. Evaluated on the FETUS 2026 leaderboard, our method achieves a Dice Similarity Coefficient at 79.99%, Normalized Surface Distance at 61.62%, and F1-score at 41.20%, validating the effectiveness of our approach for prenatal congenital heart disease screening. Source code is publicly available at the following link: https://github.com/2826056177/zcst_fetus2026.

***Index Terms*—** Semi-supervised learning, fetal cardiac ultrasound, medical image segmentation, medical image classification, multi-task learning

## 1. INTRODUCTION

Congenital heart disease (CHD) represents the most prevalent structural anomaly in fetuses and remains a leading cause of morbidity and mortality in newborns [2]. Prenatal screening for CHD relies heavily on the expert assessment of ultrasound images from standard cardiac views, including the four-chamber view (4CH), left ventricular outflow tract view (LVOT), right ventricular outflow tract view (RVOT), and three-vessel trachea view (3VT) [8,11]. However, the accurate interpretation of these views demands extensive clinical expertise and is susceptible to significant inter-observer variability [2]. Furthermore, the scarcity of large-scale, multi-view datasets with complete annotations has impeded the development of robust artificial intelligence systems capable of supporting routine prenatal assessments.

## 2. RELATED WORKS

### 2.1. Medical Image Segmentation

U-Net [9] and its variants have achieved remarkable success in medical imaging. Transformer-based models like UNETR [1] and Swin-Transformer [13] have further advanced the field through self-attention mechanisms capturing long-range dependencies. The Segment Anything Model (SAM) [5] represents a foundation model breakthrough, with SAM-Med2D specifically optimized for medical images, demonstrating superior adaptability to medical imaging characteristics.

### 2.2. Semi-Supervised Learning

SSL has gained significant attention with deep learning advancement. Mean Teacher and self-training approaches have been widely applied to semantic segmentation. UniMatch [7] extends FixMatch [6] with Unified Perturbations, incorporating both image-space and feature-space augmentations. Despite progress, error accumulation and boundary accuracy remain challenges in SSL segmentation.

### 2.3. Applications of Foundation Models in Medical Imaging

DINOv3 [10] learns general visual representations via contrastive self-supervised learning, demonstrating strong transfer to downstream tasks. SAM-Med2D [5] adapts the original SAM for medical images through continued pre-training on medical datasets, preserving segmentation capabilities while improving medical image adaptability.

* Equal Contribution
† Corresponding Author

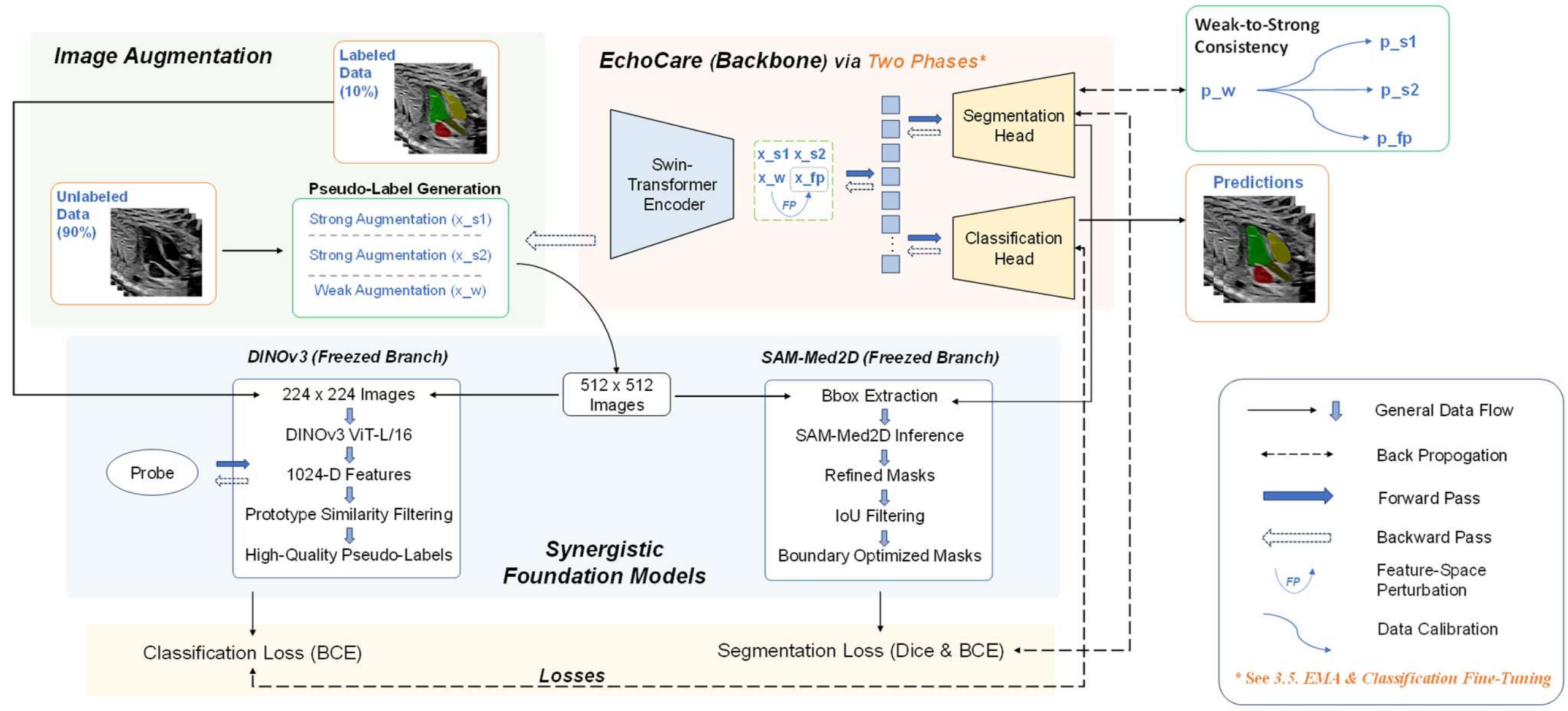

Fig. 1. Overall Framework

### 2.4. Multi-task learning

EchoCare [3] addresses joint segmentation, CHD classification, and view classification through a multi-task architecture with shared Vision Transformer encoder. However, task interference during optimization often degrades performance on individual tasks, necessitating specialized training strategies.

## 3. METHODOLOGY

Figure 1 illustrates our semi-supervised framework for joint fetal cardiac ultrasound segmentation and classification. The framework builds upon EchoCare multi-task backbone and UniMatch training paradigm. To enhance pseudo-label quality, we integrate DINOv3 Semantic Enhancement Module for reliable prototype-based filtering and SAM-Med2D Refinement Module for boundary calibration. A two-stage optimization strategy first consolidates segmentation via EMA training, then recovers classification through targeted fine-tuning without compromising segmentation performance.

### 3.1. The EchoCare Multi-Task Model

We adopt EchoCare [3] as our backbone, a multi-task architecture comprising Vision Transformer encoder, UNETR-style decoder, and classification heads. he encoder employs a Vision Transformer architecture with 12 layers, 1024 hidden dimensions, and 16 attention heads, initialized with pre-trained self-supervised weights to capture multi-scale global contextual information. The decoder outputs: 15 segmentation categories (1 background + 14 cardiac structures), 7-dimensional CHD classification via global average pooling, and 4 view types through auxiliary classification. Shared encoder enables information complementarity across tasks.

### 3.2. UniMatch

UniMatch [7] extends FixMatch [6] with weak-to-strong consistency. For unlabeled image $x_u$, weak augmentation generates pseudo-label $pw$ supervising strong augmentation prediction $p_s$. UniMatch adds Unified Perturbations (UniPerb): feature-space perturbation via channel-wise Dropout produces additional prediction $p_fp$, forming dual-path perturbation for enhanced unlabeled data exploration. Multi-task loss weighting: supervised segmentation 1.0, classification 0.8; unsupervised segmentation (S1,S2) 0.3, focal 0.4, mixed 0.2; pseudo-label classification 1.0, mixed 0.3, focal mixed 0.4.

### 3.3. Feature Extraction and Pseudo-Label Optimization via DINOv3

In semi-supervised learning, the quality of pseudo-labels is the primary bottleneck. To improve it, we employ DINOv3 [10], a SOTA self-supervised ViT, as an external semantic anchor. Unlike standard decoders, DINOv3 extracts high-level representations that are invariant to the speckle noise typical of ultrasound. We implement a prototype-based filtering mechanism where unlabeled samples are only assigned pseudo-labels for classification if their DINOv3 embeddings demonstrate high cosine similarity to established category prototypes, effectively pruning noisy supervision.

### 3.4. SAM-Med2D Assisted Pseudo-Label Refinement

The accuracy of segmentation boundaries remains a critical challenge. Pseudo-labels generated by standard methods often suffer from blurry or inaccurate boundaries, directly impacting model performance. SAM-Med2D [5], a segmentation foundation model optimized for medical images, offers a powerful solution.

We integrate SAM-Med2D into our SSL pipeline to refine the boundaries of model-generated pseudo-labels. The process is as follows: (1) an initial pseudo-label mask is generated for an unlabeled image by the current model; (2) bounding boxes are extracted from this initial mask and used as prompts for SAM-Med2D; (3) SAM-Med2D produces a refined segmentation result; (4) high-quality refined results are selected for subsequent training by comparing the Intersection over Union (IoU) between the SAM-Med2D output and the initial mask.

### 3.5. EMA for Segmentation & Fine-Tuning for Classification

Although the EMA training stage endows the model with strong segmentation capabilities, the classification performance is significantly compromised due to several factors: (1) the smoothing effect of EMA reduces the model's sensitivity to subtle classification features; (2) frequent hyperparameter tuning leads to overfitting on the validation set, causing indirect data leakage; and (3) the multi-task optimization dynamics inherently bias the model toward the segmentation task. To address this, we propose a targeted classification fine-tuning strategy that consists of two sequential phases.

Phase 1: Strong Segmentation Training with EMA. During this phase, we employ intensive segmentation training to build robust segmentation capabilities. The EMA mechanism stabilizes training and improves generalization, resulting in significant improvements in Dice and NSD metrics. However, this comes at the cost of classification performance degradation.

Phase 2: Classification Fine-tuning. In this phase, we freeze all segmentation-related parameters, including the segmentation head and the majority of the backbone encoder, to preserve the acquired segmentation performance. We then unfreeze the last semantic layer of the backbone to allow high-level semantic features to adapt to classification tasks without disrupting the low-level detail features captured by earlier layers. Meanwhile, the classification head is reset to its initial random state to eliminate any suboptimal weight configurations induced by multi-task interference. During fine-tuning, only the classification head and the unfrozen last semantic layer are trained.

Additionally, we disable strong intervention mechanisms such as SAM and mask guidance during this phase to prevent changes in segmentation pseudo-labels from interfering with classification training. It is important to note that due to multiple rounds of hyperparameter tuning, the validation set exhibits indirect data leakage, resulting in artificially elevated validation scores during the early stages of Phase 2.

To implement this strategy effectively, we employ grouped learning rates for the unfrozen last layer and the classification head. The learning rate for the last layer is set slightly lower to prevent disrupting segmentation stability, while the classification head uses a higher learning rate to accelerate convergence. Furthermore, we narrow the unfreezing scope to the last layer only, effectively controlling segmentation stability while enabling classification realignment.

## 4. EXPERIMENTAL DESIGN AND RESULT ANALYSIS

### 4.1. Dataset and Task Description

The FETUS 2026 Challenge dataset comprises 5,000 standard-view B-mode fetal cardiac ultrasound images, among which 2,800 are allocated for training (partially sourced from the FOCUS dataset [11]). Collected across multiple centers and devices, the dataset exhibits real-world characteristics including motion artifacts and speckle noise, thereby enhancing its clinical applicability.

### 4.2. Experimental Setup

Model training employs a multi-GPU parallel strategy with a batch size of 8, utilizing the AdamW optimizer (weight decay: 0.01) with differential learning rates (1e-4 for backbone, 1e-3 for task heads) over 300 epochs with polynomial decay and standard data augmentation (random flipping, rotation, scaling, color jitter). To reflect anatomical constraints, view-specific hard masking restricts segmentation categories per standard view: the four-chamber view (4CH) permits categories 0–7; the left ventricular outflow tract view (LVOT) allows categories 0, 1, 2, 4, and 8; the right ventricular outflow tract view (RVOT) includes categories 0, 6, 8–12; and the three-vessel trachea view (3VT) encompasses categories 0, 9, 12–14, thereby preventing implausible predictions and enhancing segmentation accuracy.

### 4.3 Training Strategy

The training pipeline comprises three stages designed to jointly optimize segmentation and classification while mitigating negative transfer.

Initially, a lightweight probe head is pre-trained on a frozen DINOv3 ViT-L/16 [10] backbone using labeled data to enhance pseudo-label reliability as an auxiliary filter. Subsequently, the EchoCare model undergoes joint training within the UniMatch [7] framework utilizing Exponential Moving Average (EMA) for stable optimization, initialized with pretrained encoder weights and the Stage 1 probe, while employing view-specific hard masking, SAM-Med2D boundary refinement, and DINOv3 prototype-based filtering.

Finally, to address classification performance degradation caused by EMA smoothing and multi-task interference, a classification fine-tuning strategy is implemented where segmentation parameters are frozen, the classification head is reset, and only the terminal backbone layer and classification head are trained to realign discriminative performance without compromising segmentation integrity.

### 4.4. Evaluation Metrics

Segmentation performance is evaluated using the Dice Similarity Coefficient (DSC), which measures region overlap, and the Normalized Surface Distance (NSD), which quantifies boundary discrepancy. Classification performance is assessed using the macro-averaged F1-Score across all categories. The final challenge ranking is determined by a weighted combination of the F1-Score, DSC, and NSD.

### 4.5. Experimental Results and Analysis

To verify the effectiveness of our method, we conducted comparative experiments within the UniMatch framework. The experimental results are presented in Table 1.

| | Methods | F1-Score (%) | DSC (%) | NSD (%) | Overall Score (%) |
|---|---|---|---|---|---|
| a) | EchoCare (Baseline) | 34.20 | 65.48 | 45.55 | 44.86 |
| b) | EchoCare + SAM-Med2D | 28.44 | 75.92 | 56.62 | 47.36 |
| c) | EchoCare + SAM-Med2D + DINOv3 | 39.03 | 74.71 | 54.76 | 51.88 |
| d) | EchoCare + SAM-Med2D + DINOv3 (EMA) | 25.25 | 80.04 | 61.54 | 48.02 |
| e) | EchoCare + SAM-Med2D + DINOv3 (Classification Fine-tuning based on *method d)*) | **41.20** | **79.99** | **61.62** | **56.00** |

**Table 1.** Segmentation and Classification Performance Comparison of Different Methods

Our proposed *method e)* achieves the best overall performance with an Overall Score of 56.00, improving DSC from 65.48% to 79.99% (+14.51%), NSD from 45.55% to 61.62% (+16.07%), and F1-Score from 34.20% to 41.20% (+7.00%), compared to the *baseline*.

The analysis reveals that adding SAM-Med2D boundary refinement further improves DSC by 10.44% and NSD by 11.07%, and DINOv3 effectively boosts F1-Score by 10.59%. EMA training significantly reinforces pixel-level segmentation metrics, while it leads to a severe F1-Score degradation from 39.03% to 25.25%. To address this trade-off, our classification fine-tuning strategy successfully regains classification performance from 25.25% to 41.20% while preserving the segmentation gains.

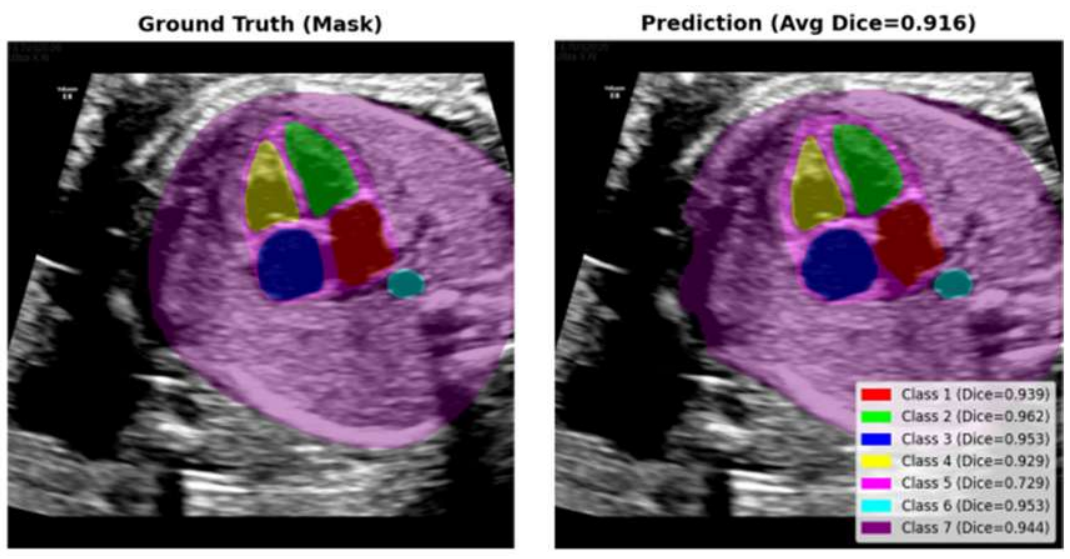

**Fig. 2.** Comparison between *Ground Truth* and *Method e)*

## 5. CONCLUSION

This paper presents a semi-supervised framework for joint fetal cardiac ultrasound image segmentation and classification, built upon the EchoCare multi-task backbone and UniMatch training paradigm. By integrating SAM-Med2D for boundary refinement and DINOv3 for semantic enhancement of pseudo-labels, our method achieves significant improvements over the baseline, with DSC, NSD, and F1-score reaching 79.99%, 61.62% and 41.20% respectively on the FETUS 2026 leaderboard.

The key contributions of this work are threefold: (1) We introduce SAM-Med2D-assisted boundary refinement into semi-supervised learning, which significantly improves DSC and NSD; (2) We leverage DINOv3 features as an external semantic anchor to filter unreliable pseudo-labels through prototype matching, enhancing the F1-Score; (3) We design a view-specific hard masking mechanism that constrains segmentation categories based on anatomical priors, effectively reducing false predictions. Furthermore, to address the classification performance degradation caused by EMA training and multi-task interference, we propose a two-stage fine-tuning strategy that freezes segmentation parameters and resets the classification head, enabling F1-score to rebound to a higher level.

Future work will focus on optimizing the method combining SAM-Med2D and DINOv3 to explore ultrasound-specific pre-training strategies, and extending the framework to other medical image analysis tasks.

## 6. COMPLIANCE WITH ETHICAL STANDARDS

This study is conducted retrospectively using human subject data made available in open access by the FETUS 2026 Challenge organizers. Ethical approval is not required as confirmed by the license attached with the open access data.

## 7. ACKNOWLEDGMENTS

This work was supported by the Guangdong Key Disciplines Project under grant number 2024ZDJS137.